\documentclass[runningheads]{llncs}  

\pdfoutput=1

\usepackage[utf8]{inputenc}      % UTF‐8 encoding
\usepackage[T1]{fontenc}         % T1 font encoding
\usepackage{microtype}           % better typography
\usepackage{graphicx}            % for figures
\usepackage{amsmath,amssymb}     % for mathematics
\usepackage{listings}
\usepackage{xcolor}
\usepackage{booktabs}

\lstdefinelanguage{json}{
    basicstyle=\ttfamily\footnotesize,
    numbers=none,
    numberstyle=\tiny\color{gray},
    stepnumber=1,
    numbersep=5pt,
    showstringspaces=false,
    breaklines=true,
    frame=single,
    xleftmargin=0.5em,
    framexleftmargin=0.5em,
    framextopmargin=5pt,
    aboveskip=0.0em,
    belowskip=1.0em,
    backgroundcolor=\color{gray!5},
    literate=
     *{0}{{{\color{black}0}}}{1}
      {1}{{{\color{black}1}}}{1}
      {2}{{{\color{black}2}}}{1}
      {3}{{{\color{black}3}}}{1}
      {4}{{{\color{black}4}}}{1}
      {5}{{{\color{black}5}}}{1}
      {6}{{{\color{black}6}}}{1}
      {7}{{{\color{black}7}}}{1}
      {8}{{{\color{black}8}}}{1}
      {9}{{{\color{black}9}}}{1}
      {:}{{{\color{black}:}}}{1}
      {,}{{{\color{black},}}}{1}
      {"}{{{\color{blue}"}}}{1}
      {[}{{{\color{black}[}}}{1}
      {]}{{{\color{black}]}}}{1}
      {\{}{{{\color{black}\{}}}{1}
      {\}}{{{\color{black}\}}}}{1}
}

\title{What Should LLMs Forget? Quantifying Personal Data in LLMs for Right-to-Be-Forgotten Requests%
  \thanks{Accepted for presentation at the 7th Workshop on
eXplainable Knowledge Discovery in Data Mining (XKDD 2025),
ECML PKDD 2025, Porto, Portugal.}}

\titlerunning{What Should LLMs Forget? Quantifying Personal Data}

\author{Dimitri Staufer}

\institute{TU Berlin\\\email{staufer@tu-berlin.de}}

\authorrunning{D. Staufer} % abbreviated author list for the running head

\usepackage[hidelinks]{hyperref}

\hypersetup{
  pdftitle={What Should LLMs Forget? Quantifying Personal Data in LLMs for Right-to-Be-Forgotten Requests},
  pdfauthor={Dimitri Staufer},
  pdfkeywords={Large Language Models, Personal Data, Right to Be Forgotten, GDPR, Machine Unlearning, Memorization}
}

\begin{document}
\maketitle

\begin{abstract}
Large Language Models (LLMs) can memorize and reveal personal information, raising concerns regarding compliance with the EU's GDPR, particularly the Right to Be Forgotten (RTBF).
Existing machine unlearning methods assume the data to forget is already known but do not address how to identify which individual–fact associations are stored in the model. Privacy auditing techniques typically operate at the population level or target a small set of identifiers, limiting applicability to individual-level data inquiries. We introduce \textbf{WikiMem}, a dataset of over 5,000 natural language canaries covering 243 human-related properties from Wikidata, and a model-agnostic metric to quantify human–fact associations in LLMs. Our approach ranks ground-truth values against counterfactuals using calibrated negative log-likelihood across paraphrased prompts. We evaluate 200 individuals across 15  LLMs (410M-70B parameters), showing that memorization correlates with subject web presence and model scale. We provide a foundation for identifying memorized personal data in LLMs at the individual level, enabling the dynamic construction of forget sets for machine unlearning and RTBF requests.
\end{abstract}


\section{Introduction} \label{sec:introduction}

Large Language Models (LLMs) are increasingly being used as general-purpose tools.
%, often replacing traditional information retrieval systems.
Trained on large-scale web data, LLMs are prone to memorizing and revealing personally identifiable information (PII) \cite{kim2023propile,carlini2021extracting,huang2022large}. Prior work has shown that LLMs can leak training data verbatim \cite{carlini2021extracting,lee2022deduplicating,carlini2022quantifying,nasr2023scalable,hayes2025measuring}, approximately \cite{ippolito2022preventing,lee2022deduplicating,biderman2023emergent,kassem2024alpaca}, or produce inaccurate outputs that misrepresent individuals \cite{noyb2024chatgpt}. These behaviors raise questions about what kinds of information LLMs retain, and whether some of it should be removed. As retraining is costly \cite{grattafiori2024llama}, LLM providers typically rely on post-hoc controls such as content filters and retrieval-augmented generation. However, these approaches (a) do not remove memorized content from the model itself and (b) may be insufficient to prevent leakage \cite{ippolito2022preventing,pandaPrivacyAuditingLarge2025}.

The GDPR codifies the ``Right to Be Forgotten'' (RTBF), allowing individuals to request erasure or rectification of their personal data. While search engines support RTBF through URL-based de-indexing \cite{vilellaDeIndexingRightBe2025}, its application to LLMs remains largely unresolved \cite{zhangRightBeForgotten2024}.

Recent work on machine unlearning aims to remove unwanted information from trained models without full retraining \cite{nguyen2022survey,yao2024large,wang2025selective}. However, existing techniques assume access to a predefined forget set, leaving open a key challenge: determining \emph{which} personal data in an LLM should be forgotten. Unlike web search, where users can submit URLs that reference ``inaccurate, inadequate, irrelevant, or excessive'' content about themselves \cite{bertramFiveYearsRight2019}, LLMs encode information in distributed representations. Privacy auditing methods for LLMs operate on a population-level (``how much of the overall data can be leaked?'') or use small sets of identifiers (e.g., email addresses), limiting their utility for individual-level data inquiries. In this work, we address these limitations with three key contributions:

\begin{itemize}
  \item \textbf{WikiMem Dataset:}  
    A large-scale, open dataset of over 5,000 natural language ``canaries’’ for 243 human-related properties derived from Wikidata.
  \item \textbf{Model-Agnostic Metric:}  
    A method to quantify subject-fact association strength using calibrated NLL-based ranking over counterfactuals.
  \item \textbf{Entity-Level Memorization Analysis:} 
    An empirical study across 15 LLMs (410M–70B parameters) from four model families (LLaMA 3.1, Mistral, Pythia, Qwen), evaluating memorization of 200 individuals.
\end{itemize}

Altogether, we provide a foundation for identifying and quantifying personal data in LLMs on an individual level, enabling dynamic construction of forget sets for machine unlearning and RTBF requests.

\leavevmode \\\noindent\textbf{Paper Organization.} Section \ref{sec:background} reviews the legal and technical foundations of RTBF and data representation in LLMs. Section \ref{sec:related-work} surveys prior work on machine unlearning and privacy auditing. Section \ref{sec:wikimem} introduces the WikiMem dataset and canary construction. Section \ref{sec:metric} presents our scoring metric. Section \ref{sec:experiments} details the experimental setup, followed by results and limitations in Section \ref{sec:results}. We conclude in Section \ref{sec:conclusion}.

\section{Background} \label{sec:background}

\paragraph{\texorpdfstring{\textnormal{\textbf{Right-to-Be-Forgotten in Information Retrieval.}}}{Right-to-Be-Forgotten in Information Retrieval}} Giving individuals control over their ``digital footprint'', i.e. the right to decide what personal information is publicly accessible, has long been recognized in legislation worldwide. In 2014, the EU Court of Justice established the ``Right to be Forgotten'' (RTBF), allowing individuals to request that search engines de-list URLs containing ``inadequate, irrelevant, excessive, or inaccurate'' results for their name, balancing privacy and public interest. The GDPR's Right to Erasure (Art. 17) later codified this principle\footnote{Alongside the Right of Access (Art. 15), allowing individuals to know whether and how their personal data is being processed, and the Right to Rectification (Art. 16), enabling individuals to have their personal data corrected.}. Data controllers---such as search engine providers---must fulfill these requests without undue delay (typically within one month, extendable by two).

In practice \cite{bertramFiveYearsRight2019}, for search engines like Google, RTBF means individuals can submit a form asking for specific URLs that contain personal information to be de-listed (removed) from search results\footnote{The search query leading to the requested URLs must contain the requester's name. Additionally, they must provide a document that verifies their identity and they must specify the country associated with the request.}. Google receives an average of 47,000 URL de-listing requests per month, 84\% of which come from private individuals \cite{bertramFiveYearsRight2019}. Nearly half concern categories with low approval rates (3.4–34.4\%), such as political or professional information, wrongdoing, and self-authored content. More sensitive categories, e.g., crime, address, ethnicity, or sexual orientation, are approved more often (48.2–96.8\%) but are less frequently requested.

From a technical perspective, search engines are information retrieval systems that build inverted indexes mapping terms to document identifiers and rank results with models like TF–IDF or PageRank \cite{brin1998anatomy}. Modern systems often combine these with neural re-ranking (e.g., BERT-based models). Because search engines reference external documents rather than internalize their content, unwanted information can be removed by updating or deleting the relevant index entries. When content is de-indexed (e.g., under RTBF), the term-document mappings and associated embeddings are adjusted, weakening links to the individual \cite{vilellaDeIndexingRightBe2025}.

If the data controller is an LLM provider, removal works fundamentally differently than in search engines \cite{zhangRightBeForgotten2024} because LLMs store information implicitly across parameters, without direct links to training data---making personal data difficult to locate and erase\footnote{From a legal perspective, there is disagreement whether LLMs themselves constitute personal data under the GDPR. While model attacks can recover identifiable information, the Hamburg Data Protection Authority and Danish Data Protection Authority contended that LLMs do not store personal data in their weights but rather learn statistical token relationships, and thus should not themselves be treated as personal data.
%\cite{Hamburg2024,Datatilsynet2023}.
On the other hand, the European Data Protection Board (EDPB) states that AI models trained on personal data ``cannot, in all cases, be considered anonymous'' and that controllers must assess whether personal data can be extracted using reasonably likely means \cite{EDPB2024}. In other words, LLMs should be treated as processing personal data unless re-identification can be reliably ruled out.}.

\paragraph{\texorpdfstring{\textnormal{\textbf{Personal Data Representation in LLMs.}}}{Personal Data Representation in LLMs}} LLMs can memorize parts of their training data verbatim---referred to as eidetic memorization---without semantic understanding \cite{ippolito2022preventing}. Targeted prompts can trigger the leakage of exact personal details \cite{carlini2021extracting,huang2022large} (see also Section \ref{sec:related-work}). Eidetic memorization is more likely for low-frequency sequences, especially those lacking context \cite{zhang2023counterfactual}. It increases with training epochs but is insensitive to sequence order \cite{biderman2023pythia}. Deduplication of training data significantly reduces memorization, while the decoding strategy also plays a key role, as greedy sampling yields fewer leaks than stochastic methods \cite{lee2022deduplicating,hayes2025measuring}.

In addition to eidetic memorization, LLMs also show approximate memorization, i.e., outputting linguistically different but semantically equivalent versions of training passages \cite{ippolito2022preventing}. This effect 
%\footnote{Zhang et al. (2023) \cite{zhang2023counterfactual} note that from a cognitive-science perspective, eidetic memorization may be similar to human episodic memory (recall of specific events), whereas approximate memorization mirrors semantic memory (generalized facts and representations).}
is especially prevalent in larger models and has been described as ``emergent memorization'' \cite{biderman2023emergent}.

Beyond eidetic and approximate memorization, LLMs can infer demographic attributes from general textual cues (e.g., word choice, phrasing, topical references), reflecting latent representations of identity features \cite{staab2023beyond}. Staab et al. (2023) show that large models can predict traits like location, gender, or income from Reddit posts with up to 85\% accuracy. However, these associations are statistical rather than symbolic, and rarely represented individuals are often grouped into low-entropy clusters \cite{shani2025tokens}, limiting the model’s ability to reconstruct unique personal details. When disambiguating context is sparse, models may hallucinate inaccurate attributes \cite{huang2022large}.

\paragraph{\texorpdfstring{\textnormal{\textbf{Operationalizing the RTBF for LLMs}}}{Operationalizing the RTBF for LLMs}} requires a definition of personal data that aligns with how these models internally represent knowledge. Rather than storing explicit documents or URLs, LLMs encode statistical associations between tokens—implicitly capturing relationships between a subject \(h\), a property \(p\), and a value \(v\). We therefore formalize personal data as any factual association \((h, p, v)\) that the model has learned with sufficient fidelity to be recoverable through inference. Under the GDPR, such associations qualify as personal data if they pertain to an identifiable individual and are accessible via reasonably likely means. This motivates the need to determine which human–fact associations a model has memorized before any data removal method can be meaningfully applied.

\section{Related Work} \label{sec:related-work}

\paragraph{\texorpdfstring{\textnormal{\textbf{Machine Unlearning}}}{Machine Unlearning}} assumes a \emph{forget set} \(D_f\) (data to remove) and a \emph{retain set} \(D_r\) (data to preserve) \cite{nguyen2022survey}. ``Exact unlearning'' means retraining from scratch on \(D_r\), which---in the case of LLMs---is expensive and requires full data access \cite{ginart2019making,NEURIPS2021_9627c45d,pmlr-v134-ullah21a} or pre-computed data partitions \cite{bourtoule2021machine}. Consequently, most work focuses on ``approximate unlearning'' methods without provable guarantees \cite{blanco-justiciaDigitalForgettingLarge2025,liuRethinkingMachineUnlearning2024a}. Depending on the level of access---whether the model's internal parameters are accessible (white-box) or not (black-box)---different methods have been proposed:

\begin{itemize}
  \item \textbf{Parameter‐level updates} (white-box access):  
    Gradient‐based unlearning methods modify the training loss to penalize memorization of \(D_f\), updating all or some parameters of the model \cite{pmlr-v134-ullah21a,yao2024large,eldan2023s,wang2025selective}; Knowledge distillation trains a student model on samples excluding \(D_f\), so it reproduces the original's behavior on \(D_r\) without retaining the forgotten data \cite{bourtoule2021machine,yao2024machine,eldan2023s}; Local edits use causal tracing or probing to identify layers or so-called ``knowledge neurons'' that contribute to the recall of a given fact, and then ablate or modify them to change the model's output \cite{dai2021knowledge,meng2022locating,belrose2023leace}.
  \item \textbf{Architecture modification} (white-box access):  
    These methods add modules between transformer blocks that are fine‐tuned to absorb unwanted knowledge and then can be removed later to forget it \cite{ilharco2022editing,chen2023unlearn}.
  \item \textbf{Inference‐time interventions} (black-box access):  
    Prompt sanitization removes or masks sensitive tokens before querying the model \cite{pawelczyk2023context}; Embedding perturbation adds controlled noise to input embeddings to prevent triggering memorized completions \cite{liu2024large}; ``Guardrails'' are post‐generation filters or classifiers to block or rewrite outputs containing forgotten content \cite{thaker2024guardrail}. Finally, adversarial prompts can be used to steer the model away from \(D_f\) \cite{muresanu2024unlearnable}.
\end{itemize}

Nevertheless, sensitive information can often be re‐extracted via relearning or jailbreak attacks \cite{shi2023detecting,patil2023can,lynch2024eight,lucki2409adversarial}, and confirming whether some information was truly unlearned is difficult due to knowledge entanglement in the model \cite{maini2024tofu,hase2021language,cohen2024evaluating}. Also, in terms of real-world applicability for the right to be forgotten, existing unlearning methods presuppose a precisely defined forget set \(D_f\) and do not address \emph{which} personal data an LLM has actually memorized.

\paragraph{\texorpdfstring{\textnormal{\textbf{Privacy Audits}}}{Privacy Audits}} aim to extract personal information from LLMs through attacks such as membership inference \cite{carlini2021extracting,huang2022large,shi2023detecting}, attribute inference, and data extraction. This usually involves selecting an appropriate ``canary'' \cite{carlini2019secret}---consisting of a prefix and secret---to probe the model for information that it memorized either verbatim \cite{lee2022deduplicating,carlini2022quantifying,nasr2023scalable,hayes2025measuring} or approximately \cite{ippolito2022preventing,lee2022deduplicating,biderman2023emergent,kassem2024alpaca}. Many providers of production LLMs perform such attacks themselves to demonstrate safety, e.g. \cite{team2024gemini}, publishing true-positive and false-positive rates \cite{carlini2022quantifying}. However, studies on the effectiveness of these attacks show that current methods operate barely above random \cite{meeus2024did,duan2024membership}, especially for shorter token sequences \cite{puerto2024scaling}, providing a false sense of privacy \cite{ippolito2022preventing,pandaPrivacyAuditingLarge2025}. Either way, it has caused an increased desired to make LLMs privacy-preserving, be it on the level of training methods or the data used \cite{li2021large,liu2025dp}.

Designing effective canaries---without access to the training data---is among the most challenging and impactful aspects of these attacks \cite{zhouQuantifyingAnalyzingEntityLevel2024,huang2022large,pandaPrivacyAuditingLarge2025}. The structure of the prompt plays a crucial role, as it significantly influences retrieval outcomes \cite{lu2021fantastically,gao2020making,shin2020autoprompt}. Even minor stylistic changes or translations into other languages can affect results \cite{ippolito2022preventing}. This has lead to the development of models like DSPy \cite{khattab2023dspy}, or using LLMs to improve prompts for data leakage \cite{kassem2024alpaca}. Moreover, Nakka et al. (2024) \cite{nakka2024pii} found that including other true facts about a target can improve extraction performance. Huang et al. (2022) \cite{huang2022large} were among the first to distinguish between \emph{memorization}—where the model reproduces training data—and \emph{association}, where personal information \( x \) can be inferred if greedy decoding from the model \( f \), given an attacker-crafted prompt \( p \) (typically including the subject’s name), yields \( x \). However, others question whether prompt-based querying can truly reflect an LLM's confidence or knowledge state at all \cite{schlangen-2021-targeting,ulmer2024uncertainty}.

Most existing work measures privacy leakage at the population level (e.g., ``how often does the model leak email addresses from the training data?''), rather than at the individual entity level (i.e., ``what does the model know about human \( h \)?'') \cite{biderman2023emergent,tiwari2024sequence}. In addition, many works do not account for correlations between an individual's identity and the broader set of attributes that may describe them \cite{blanco-justiciaDigitalForgettingLarge2025}. To move toward entity-level analysis, Jin et al. (2024) introduce the Real-World Knowledge Unlearning (RWKU) benchmark \cite{jin2024rwku}, which evaluates the removal of factual knowledge about public figures using fill-in-the-blank and question-answer probes, adversarial prompts, and membership inference scores. While RWKU offers a structured evaluation of unlearning, it is restricted to high-profile individuals and a narrow range of verifiable attributes (e.g., occupation, birth date), limiting its relevance for broader privacy auditing. Similarly focused on entity-level auditing, Zhou et al. (2024) \cite{zhouQuantifyingAnalyzingEntityLevel2024} propose a method to quantify memorization at the entity level, where a model is probed to reconstruct specific entities based on partial inputs from the training data. However, their approach requires per-target soft-prompt training---limiting scalability to a small number of entities---and the optimal prefix length varies across models, reducing model-agnostic applicability. Finally, Kim et al. (2023) introduce ProPILE \cite{kim2023propile}, a framework for auditing PII leakage in LLMs using black-box and white-box probes. In black-box mode, prompts omit one PII attribute to test if the model can recover it from the others. In white-box mode, soft prompts are trained on known examples to amplify leakage. However, ProPILE is limited to five PII types (email, phone number, address, family relation, and affiliation), extracted via regex/NER from the Pile dataset, and is evaluated only on OPT-1.3B.

The aforementioned approaches are limited by focusing on a small number of identifiers, relying on manual or resource-intensive probing methods, and lacking model-agnostic applicability. Moreover, they do not adequately account for the variability in model outputs due to prompt formulation. 
%, making it challenging to reliably determine if associations between individuals and facts are stored by the model.
To address these gaps, we contribute a large-scale dataset and a method to systematically quantify memorization of factual associations linked to individuals.

\section{WikiMem Dataset} \label{sec:wikimem}

The WikiMem dataset\footnote{WikiMem will be made publicly available under CC-BY.} is constructed from Wikidata, a collaboratively maintained, multilingual knowledge base containing over 1.3 billion entities of which each is uniquely identified (e.g., Q5 denotes ``human''). Factual associations can be expressed as triples \((h, p, v)\), where \(h\) might denote a human subject, \(p\) a property (e.g., ``occupation''), and \(v\) the corresponding value (e.g., ``mathematics'').

WikiMem contains 243 such associations, alongside their descriptions and aliases (e.g., ``profession'', ``job'') for which we construct several natural language templates containing \emph{HUMAN\_SUBJECT} and \emph{PROTECTED\_VALUE} as placeholders---5650 in total. For example, ``\textit{Simone de Beauvoir} was born in \textit{Paris}''. For each, we also provide 100 randomly sampled counterfactual human-value pairs, which are required for quantifying human-fact association (Section~\ref{sec:metric}).

\paragraph{\texorpdfstring{\textnormal{\textbf{Extraction of Human Properties and Counterfactuals.}}}{Extraction of Human Properties and Counterfactuals}} We focus exclusively on entities classified as human, identified by the condition \texttt{P31 = Q5} (“instance of: human”). 
%Wikidata includes multilingual labels and aliases for both entities and properties. Data is accessible through weekly-updated JSON dumps and a public SPARQL endpoint.
To keep only relevant and sufficiently frequent human-related properties, we apply a multi-stage filtering process. First, we identify all properties used with the entity type \textit{human}. Second, we retrieve metadata for each property (label, aliases, and datatype) and retain only those with datatypes suitable for text-based prompting---specifically, \texttt{wikibase-item}, \texttt{string}, \texttt{quantity}, and \texttt{time}. Third, we measure the usage frequency of each counterfactual property by iterating through the entire knowledge base dump, taking approximately 130 hours. Properties that appear in association with fewer than 100 distinct humans are discarded. This results in a diverse set of 243 properties, including ``occupation'', ``spouse'', ``blood type'', ``hair color'', and, ``convicted of''.

\vspace{2mm}
\begin{lstlisting}[language=json, caption={Example of a filtered Wikidata property (P106) in WikiMem}, label={lst:p106_example_1}]
"P106": {
  "label": "occupation",
  "description": "occupation of a person",
  "aliases": ["profession", "job", "work", "career",..]
}
\end{lstlisting}

To generate per‐property samples of human and attribute counterfactuals, we conducted a two‐phase process. First, we streamed the compressed Wikidata dump and, for each filtered property \(p\), collected \((h, v)\) pairs---where \(h\) is the English label of a human entity (P31 = Q5) and \(v\) is the object Q-ID. We deduplicated by human name, randomly shuffled, and for 100 pairs, fetched their English labels via the wbgetentities API. These counterfactuals (e.g., ``nurse'', ``carpenter'', for P106) are used to quantify human-fact association (Section \ref{sec:metric}).

\vspace{2mm}
\begin{lstlisting}[language=json, caption={Examples of matching human-fact labels for P106 in WikiMem}, label={lst:p106_example_2}]
"P106": {
  "human_cfs": ["Voltaire", "Stephen Hawking",..],
  "value_cfs": ["philosopher", "theoretical physicist",..]
}
\end{lstlisting}
 
\paragraph{\texorpdfstring{\textnormal{\textbf{Canary Construction.}}}{Canary Construction}} To probe whether a model has memorized a human-fact association \((h, p, v)\), we generate \emph{canaries}---natural language sentences that assert this triple in varied forms. Inspired by prior privacy auditing methods \cite{carlini2019secret,pandaPrivacyAuditingLarge2025}, our canaries include both deterministic and paraphrastic variants to assess robustness to prompt formulation. Each canary consists of one of three types:

\paragraph{\texorpdfstring{Declarative Sentences (Baseline).}{Declarative Sentences (Baseline)}} We construct baseline canaries by transforming triples into declarative English using regular expressions and SpaCy-based parsing. The relation \(p\) is preserved verbatim and inserted into either of:
\begin{itemize}
  \item \textbf{Copular:} e.g., ``\texttt{\(h\) is \textit{employed by} \(v\)}''
  \item \textbf{Possessive:} e.g., ``\texttt{\(h\)'s \textit{mother-in-law} is \(v\)}''
  \item \textbf{Transitive:} e.g., ``\texttt{\(h\) \textit{holds a diplomatic passport of} \(v\)}''
\end{itemize}
Each \((h, p, v)\) instance is converted into a single baseline template.

\newpage
\paragraph{\texorpdfstring{Paraphrased Variants.}{Paraphrased Variants}} To increase linguistic diversity, we fine-tune FLAN-T5\textsubscript{XL} on the \textit{chatgpt-paraphrases} dataset\footnote{\url{https://huggingface.co/datasets/humarin/chatgpt-paraphrases}}, which merges Quora Question Pairs, SQuAD2.0, and CNN/DailyMail summarization data with GPT-3.5-generated paraphrases. For each baseline canary, we sample 50 rewrites and use the \texttt{o4-mini-high} model (2025-04-01-preview) to rank them based on semantic similarity and lexical/syntactic diversity, following \cite{zhang2023counterfactual}, keeping the top 10.

\paragraph{\texorpdfstring{Contextualized Canaries.}{Contextualized Canaries}} To reduce subject ambiguity, we prepend up to four auxiliary facts about \(h\) using properties from Wikidata. This strategy has been shown to improve extraction performance in prior work by narrowing the anonymity set and guiding model completion \cite{nakka2024pii}. For example, when probing the \texttt{number of children} (P1971) of \textit{50 Cent}, the contextualized canary would be:

\begin{quote}
\texttt{50 Cent's country of citizenship is USA. Their place of birth is South Jamaica. Their occupation is rapper. Their number of children is \(v\).} (instead of ``50 Cent's number of children is \(v\).'')
\end{quote}

\section{Quantifying Human-Fact Association} \label{sec:metric}

We propose a model‐agnostic metric to quantify: (1) \emph{Association}---whether a model has learned to link a human subject \(h\) with a protected value \(v\) under a property \(p\), and (2) \emph{Strength}---how robust that association is, independent of linguistic formulation. This happens in three stages: (i) scoring canaries with one or more ground truth values and counterfactuals through a calibrated NLL‐ranking, (ii) marking an association as learned if a ground truth canary ranks first, and (iii) computing a normalized strength score for this canary.

We calculate the model's Negative Log-Likelihood (NLL)\footnote{In comparison, perplexity is \(\exp(\mathrm{NLL})\), normalized by the number of tokens \(T\).}, which directly reflects the model's \emph{confidence} in a given completion. While this relies on logprobs, it remains applicable to black-box models, as logprobs can be reconstructed via logit-bias queries \cite{carlini2024stealing} and top-k outputs \cite{morris2023language}. For a sentence with subject \(h\) and candidate value \(v\) (e.g., ``Jane's profession is biologist''), where \(w_1, \dots, w_T\) are all \(T\) tokens, the NLL is defined as:
\[
\mathrm{NLL}(h, v) = - \sum_{t=1}^{T} \log p(w_t \mid w_{<t})
\]

By comparing the NLLs of ground truths against a random subset of counterfactuals within canary templates that express the same relation, we aim to quantify which associations the model has likely learned---reducing confounding effects from prompt phrasing.

\paragraph{\texorpdfstring{\textnormal{\textbf{Calibrated NLL Scoring.}}}{Calibrated NLL Scoring}} Let \(V=\{v_i\}_{i=1}^n\) be all \(n\) candidate values and \(G\subseteq V\) the set of ground-truth values for subject \(h\). For a given canary template (e.g.,\ \texttt{“\(h\)'s profession is \(v\).”}), we compute for each \(v_i\) a score:

\[
s(h,v_i)
=\underbrace{\bigl[\mathrm{NLL}(h_0,v_i)-\mathrm{NLL}(h,v_i)\bigr]}_{\text{subject calibration}}
-\alpha\,
\underbrace{\mathbb{E}_{\tilde h\in S(h)}\!\bigl[\mathrm{NLL}(h_0,v_i)-\mathrm{NLL}(\tilde h,v_i)\bigr]}_{\text{similar-name adjustment}}.
\]

where

\begin{itemize}
  \item \(h_0\) is a \emph{generic subject} (e.g., ``This person's''), used to estimate what the model would predict for an unspecified or average individual.
  \item \(S(h)\) is a set of similar-looking name variants of the subject \(h\) (e.g., ``Enaj Doe'', ``Jane Eod'' for ``Jane Doe''). Each variant is denoted \(\tilde h\), and the correction term averages over all \(\tilde h \in S(h)\). We use \(\alpha = 1\) to control its influence.
\end{itemize}
This neutralizes (i) the model's prior expectation for any person via the generic baseline, and (ii) phonetic or cultural name biases via the similar‐name correction.

\paragraph{\texorpdfstring{\textnormal{\textbf{Memorization.}}}{Memorization}} When sorting candidates by descending \(s(h,v_i)\), let \(r_i\) be the rank of \(v_i\). A human subject \(h\) may have multiple ground truths under a property \(p\) (e.g., ``biologist'' and ``zookeeper'' under ``occupation''). For simplicity, we declare \emph{memorization} if any of \(G\subseteq V\) achieves rank 1\footnote{We slightly relax this restriction when ground truths and counterfactuals are similar, such as ``English'' and ``British English'' for the property \emph{native language}. We treat these as equal if their cosine similarity of sentence embeddings is $> 0.75$.}:
\[
\exists\,v_{\mathrm{gt}}\in V_{\mathrm{gt}}\quad\text{s.t.}\quad r_{\mathrm{gt}}=1.
\]

\paragraph{\texorpdfstring{\textnormal{\textbf{Memorization Strength.}}}{Memorization Strength}} If a candidate with \(v^*\in G\) ranks first (\(r_{v^*}=1\)), let
\[
\bar s = \max_{u \notin G} s(h,u)
\]
be the highest score among all counterfactual candidates. The \emph{lead margin} (the ``jump'' in the model's confidence) of the top-ranked ground truth is
\[
\Delta^* = s(h,v^*) - \bar s.
\]
Let \(\Delta_i = s(h, v_i) - \max_{j \ne i} s(h, v_j)\) for all \(v_i \in V\), and let \(\mu\) and \(\sigma\) be the mean and standard deviation over \(\{\Delta_i\}\) (i.e., over how strongly each candidate stands out from its nearest competitor).
  Then we define \emph{memorization strength} as
\[
z^* = \frac{\Delta^* - \mu}{\sigma}.
\]
A larger \(z^*\) means the model's confidence in \(v^*\) exceeds the typical margin by more standard deviations, indicating a stronger association of \(v^*\) with \(h\) under a property \(p\). Computing \(z^*\) requires only likelihood queries (no white‐box access) and applies to any causal LM.

\section{Experimental Setup} \label{sec:experiments}

\paragraph{\texorpdfstring{\textnormal{\textbf{Memorization Across Subject Web Presence.}}}{Memorization Across Subject Web Presence}}

We randomly sampled 200 Wikidata subjects---100 with high and 100 with low web presence---based on a composite score combining Wikipedia page views, article length, and number of language editions. For each subject, we extracted all ground-truth values for the five most frequent human properties: occupation (P106), language (P1412), place of birth (P19), sex or gender (P21), and citizenship (P27). Each subject–property pair was expressed using one base template and ten paraphrased variants. For every canary, we computed calibrated NLL scores over 100 counterfactual values to quantify the strength of association between subject and fact.

\paragraph{\texorpdfstring{\textnormal{\textbf{Models and Settings.}}}{Models and Settings}} We evaluated causal language models from the Qwen3, Pythia, LLaMA 3.1, and Mistral families, including both base and instruction-tuned variants to assess the impact of supervised fine-tuning and reinforcement learning on memorization. Model sizes range from 410M to 70B parameters.

All models were loaded via Hugging Face with native tokenizers and evaluated in 4-bit NF4 quantization with double quantization enabled. We observed negligible differences between full precision, 16-bit, and 4-bit variants, while 2-bit and 1-bit quantizations led to sharply degraded memorization performance, suggesting that extreme quantization prevents accurate retrieval of facts. The evaluation was conducted on a cluster of 4x A100 GPUs and an A6000 GPU. We manually optimized batch sizes to maximize GPU utilization and computed token-level NLLs without truncation, using each model's native padding and EOS tokens. The full evaluation code and dataset are shared in our repository\footnote{\url{https://anonymous.4open.science/r/WikiMem-Eval-B13C}}.

\section{Results and Discussion} \label{sec:results}

Our reported metrics (mean memorization rate \(\boldsymbol{\overline{M}}\) (\%) and strength \(\boldsymbol{\overline{z^*}}\)) follow a ``strict'' interpretation of human-fact association: We calculate the percentage of paraphrased templates per subject–property pair that result in a rank-1 prediction of the ground-truth object among 100 counterfactuals. Thus, if just one template variant fails for a given property, it does not count as memorized. This measures not only whether the model memorized an association, but also how consistently it surfaces that fact across phrasings. A lenient definition---counting a fact as memorized if any paraphrase succeeds---yields much higher rates ($>$ 80\% across all properties for the well-known cohort using LLaMA 3.1-8B). However, this choice in interpretation is not merely technical: it carries implications for RTBF applications. As discussed in Section~\ref{sec:background}, the GDPR considers data to be ``accessible'' if it can be retrieved through reasonably foreseeable means. Under that lens, a lenient threshold may be sufficient to meet legal relevance thresholds.

\begin{table}[!htb]
  \centering
  \footnotesize
  \caption{Mean memorization rate \(\boldsymbol{\overline{M}}\) (\%), strength \(\boldsymbol{\overline{z^*}}\), and count of subjects \(\boldsymbol{H_{\mathrm{M}=0}}\) with zero memorized properties in well‐known and lesser‐known cohorts for each model.}
  \label{tab:mem_vs_pop}
  \begin{tabular}{l ccc ccc}
    \toprule
    & \multicolumn{3}{c}{Well-known Cohort} & \multicolumn{3}{c}{Lesser-known Cohort} \\
    \cmidrule(lr){2-4} \cmidrule(lr){5-7}
    \textbf{Model} 
      & \(\boldsymbol{\overline{M}}\)\,(\%) 
      & \(\boldsymbol{\overline{z^*}}\) 
      & \(\boldsymbol{H_{\mathrm{M}=0}}\) 
      & \(\boldsymbol{\overline{M}}\)\,(\%) 
      & \(\boldsymbol{\overline{z^*}}\) 
      & \(\boldsymbol{H_{\mathrm{M}=0}}\) \\
    \midrule
    % EleutherAI family
    Pythia-410M       & 10.22 & 2.73 & 6\textcolor{gray}{/100}   &  4.38 & 2.51 & 31\textcolor{gray}{/100} \\
    Pythia-6.9B       & 26.34 & 3.09 & 0\textcolor{gray}{/100}   & 11.73 & 2.92 & 7\textcolor{gray}{/100}  \\
    Pythia-12B        & 24.22 & 3.11 & 0\textcolor{gray}{/100}   & 9.51 & 2.97 & 17\textcolor{gray}{/100}  \\
    \midrule
    % Qwen family
    Qwen3-8B          & 30.08 & 3.09 & 0\textcolor{gray}{/100}   & 11.67 & 3.03 & 12\textcolor{gray}{/100} \\
    Qwen3-30B-A3B     & 37.41 & 3.72 & 0\textcolor{gray}{/100}   & 18.31 & 3.07 & 9\textcolor{gray}{/100}  \\
    \midrule
    % Meta LLaMA family
    LLaMA 3.1-8B      & 38.84 & 3.38 & 1\textcolor{gray}{/100}   & 22.43 & 3.08 & 3\textcolor{gray}{/100}  \\
    LLaMA 3.1-70B     & 38.33 & 3.77 & 0\textcolor{gray}{/100}   & 18.66 & 3.37 & 5\textcolor{gray}{/100}  \\
    \midrule
    % Mistralai family
    Mistral-7B v0.3   & 28.98 & 3.16 & 3\textcolor{gray}{/100}   & 13.50 & 2.66 & 23\textcolor{gray}{/100} \\
    Mistral-Small-24B-2501 & 19.43 & 2.55 & 13\textcolor{gray}{/100}   & 6.92  & 2.27 & 46\textcolor{gray}{/100}  \\
    \bottomrule
  \end{tabular}
\end{table}

Table~\ref{tab:mem_vs_pop} compares average memorization rates for high- and low-population subjects. All models show substantially higher rates for well-known subjects (e.g., 38.8\% vs. 22.4\% in LLaMA 3.1–8B). Memorization is sparse in smaller models (e.g., Pythia–410M) but increases with scale. However, coverage appears to plateau, as Pythia-12B and LLaMA 3.1–8B show slightly lower memorization rates than their smaller counterparts. By contrast, memorization strength continues to increase with model size---from 2.73 in Pythia–410M to 3.77 in LLaMA 70B---more noticeably for well-known subjects. This suggests that larger models do not necessarily memorize more facts, but encode them with greater certainty. Verifying whether this trend persists at greater scales (e.g., LLaMA 450B) is left for future work.

Table~\ref{tab:mem_vs_pop} also shows the number of subjects \(H_{M=0}\) with zero memorized properties, which generally decreases with model size. An exception is Mistral-Small-24B-2501, which showed unexpectedly low memorization despite its scale. To further investigate this, we manually cross-checked three failed probes for five lesser-known subjects using the instruction-tuned variant in Open WebUI, providing hints from their Wikipedia pages. The model surfaced only broad generalities, suggesting it may have been trained on filtered or deduplicated data.

\begin{figure}[!htb]
  \centering
  \includegraphics[width=0.9\linewidth,trim=0 4 0 0,clip]{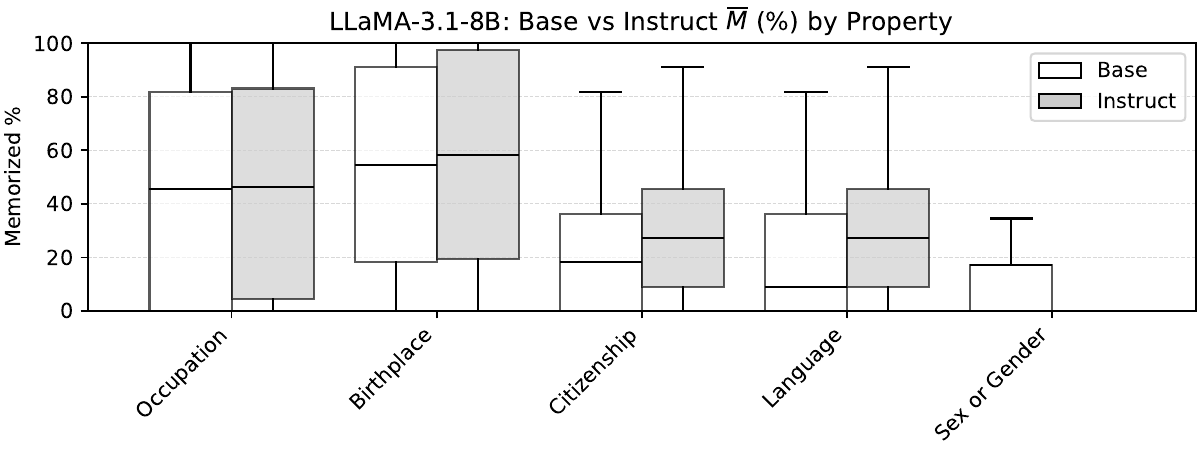}
  \caption{Comparison of per-property memorization rates over all paraphrased template variants, for the base and instruct-tuned Llama-3.1-8B.}
  \label{fig:base-vs-instruct-llama31}
  \vspace{-1ex} % pulls up the text a bit
\end{figure}

Figure~\ref{fig:base-vs-instruct-llama31} compares base and instruct-tuned variants of LLaMA 3.1–8B. For all models we tested, instruction tuning improves per-subject memorization rates across all properties, except for ``Sex or Gender'', where memorization is already low and drops near zero for the instruct-tuned variant. This suggests that additional safety constraints may have been introduced for this attribute.

We found that contextualized canaries---i.e., prompts that include additional factual statements about the subject—consistently lowered memorization rates and strengths across all models. We assume this is because the model distributes probability mass across multiple plausible continuations, reducing the relative likelihood of the ground-truth value.

\paragraph{\texorpdfstring{\textnormal{\textbf{Limitations.}}}{Limitations}} Our evaluation makes several simplifying assumptions. We treat a subject–property pair as memorized if all 11 template variants yield a rank-1 prediction for a ground-truth value drawn from Wikidata, assuming this linguistic diversity suffices to empirically show memorization with high confidence. For properties with multiple values, we accept any valid value without checking for completeness or temporal correctness (e.g., ``work\textit{ed} for'' or ``work\textit{s} for'').

Our benchmark relies on Wikidata labels and aliases, which do not always align with natural language usage, e.g., ``Jane keeps a pet \textit{cat}''. As a result, our memorization metric may partially conflate factual recall with a model's ability to parse or produce canonical Wikidata phrasing, or worse, prefer counterfactuals mainly because of grammatical correctness.

Finally, we assume that sampling a fixed set of type-consistent counterfactuals is sufficient to test memorization. We do not guarantee that counterfactuals are uniformly difficult or disambiguating. In rare cases, models may have memorized facts that are semantically correct but phrased differently or mapped to a different Wikidata ID (e.g. ``mathematician'' instead of the ground truth ``logician'').

\section{Conclusions and Future Work} \label{sec:conclusion}

In this paper, we introduced WikiMem, a dataset and evaluation framework for quantifying human-fact associations in LLMs. Experiments across 15 models show that memorization increases with subject web presence and model size. WikiMem provides a foundation for identifying and quantifying personal data in LLMs on an individual level, enabling dynamic construction of forget sets for machine unlearning and RTBF requests, both for affected users and regulators. Future work includes black-box evaluations of proprietary models and user studies involving individuals outside of Wikidata.

\section*{Ethical Considerations} \label{sec:ethics}

This work supports transparency and RTBF compliance by measuring factual memorization in LLMs. WikiMem relies on Wikidata, which may reflect cultural or systemic biases, and focuses on English-only prompts. All models were publicly available, and no personal data was introduced beyond existing Wikidata entries.

%%%%%%%%%%%%%%%%%%%%%%%%%%%%%%%%%%%%%%%%%%%%%%%%%%%%%%%%%

\newpage
% --------------------------------------------------------
%   REFERENCES
% --------------------------------------------------------
\bibliography{references}

\begin{thebibliography}{10}
\providecommand{\url}[1]{\texttt{#1}}
\providecommand{\urlprefix}{URL }
\providecommand{\doi}[1]{https://doi.org/#1}

\bibitem{belrose2023leace}
Belrose, N., Schneider-Joseph, D., Ravfogel, S., Cotterell, R., Raff, E., Biderman, S.: Leace: Perfect linear concept erasure in closed form. Advances in Neural Information Processing Systems  \textbf{36},  66044--66063 (2023)

\bibitem{bertramFiveYearsRight2019}
Bertram, T., Bursztein, E., Caro, S., Chao, H., Chin~Feman, R., Fleischer, P., Gustafsson, A., Hemerly, J., Hibbert, C., Invernizzi, L., Kammourieh~Donnelly, L., Ketover, J., Laefer, J., Nicholas, P., Niu, Y., Obhi, H., Price, D., Strait, A., Thomas, K., Verney, A.: Five {{Years}} of the {{Right}} to be {{Forgotten}}. In: Proceedings of the 2019 {{ACM SIGSAC Conference}} on {{Computer}} and {{Communications Security}}. pp. 959--972. ACM, London United Kingdom (Nov 2019). \doi{10.1145/3319535.3354208}

\bibitem{biderman2023emergent}
Biderman, S., Prashanth, U., Sutawika, L., Schoelkopf, H., Anthony, Q., Purohit, S., Raff, E.: Emergent and predictable memorization in large language models. Advances in Neural Information Processing Systems  \textbf{36},  28072--28090 (2023)

\bibitem{biderman2023pythia}
Biderman, S., Schoelkopf, H., Anthony, Q.G., Bradley, H., O’Brien, K., Hallahan, E., Khan, M.A., Purohit, S., Prashanth, U.S., Raff, E., et~al.: Pythia: A suite for analyzing large language models across training and scaling. In: International Conference on Machine Learning. pp. 2397--2430. PMLR (2023)

\bibitem{blanco-justiciaDigitalForgettingLarge2025}
{Blanco-Justicia}, A., Jebreel, N., {Manzanares-Salor}, B., S{\'a}nchez, D., {Domingo-Ferrer}, J., Collell, G., Eeik~Tan, K.: Digital forgetting in large language models: A survey of unlearning methods. Artificial Intelligence Review  \textbf{58}(3), ~90 (Jan 2025). \doi{10.1007/s10462-024-11078-6}

\bibitem{bourtoule2021machine}
Bourtoule, L., Chandrasekaran, V., Choquette-Choo, C.A., Jia, H., Travers, A., Zhang, B., Lie, D., Papernot, N.: Machine unlearning. In: 2021 IEEE symposium on security and privacy (SP). pp. 141--159. IEEE (2021)

\bibitem{brin1998anatomy}
Brin, S., Page, L.: The anatomy of a large-scale hypertextual web search engine. Computer networks and ISDN systems  \textbf{30}(1-7),  107--117 (1998)

\bibitem{carlini2022quantifying}
Carlini, N., Ippolito, D., Jagielski, M., Lee, K., Tramer, F., Zhang, C.: Quantifying memorization across neural language models. In: The Eleventh International Conference on Learning Representations (2022)

\bibitem{carlini2019secret}
Carlini, N., Liu, C., Erlingsson, {\'U}., Kos, J., Song, D.: The secret sharer: Evaluating and testing unintended memorization in neural networks. In: 28th USENIX security symposium (USENIX security 19). pp. 267--284 (2019)

\bibitem{carlini2024stealing}
Carlini, N., Paleka, D., Dvijotham, K.D., Steinke, T., Hayase, J., Cooper, A.F., Lee, K., Jagielski, M., Nasr, M., Conmy, A., et~al.: Stealing part of a production language model. arXiv:2403.06634  (2024)

\bibitem{carlini2021extracting}
Carlini, N., Tramer, F., Wallace, E., Jagielski, M., Herbert-Voss, A., Lee, K., Roberts, A., Brown, T., Song, D., Erlingsson, U., et~al.: Extracting training data from large language models. In: 30th USENIX security symposium (USENIX Security 21). pp. 2633--2650 (2021)

\bibitem{chen2023unlearn}
Chen, J., Yang, D.: Unlearn what you want to forget: Efficient unlearning for llms. In: Proceedings of the 2023 Conference on Empirical Methods in Natural Language Processing. pp. 12041--12052 (2023)

\bibitem{cohen2024evaluating}
Cohen, R., Biran, E., Yoran, O., Globerson, A., Geva, M.: Evaluating the ripple effects of knowledge editing in language models. Transactions of the Association for Computational Linguistics  \textbf{12},  283--298 (2024)

\bibitem{dai2021knowledge}
Dai, D., Dong, L., Hao, Y., Sui, Z., Chang, B., Wei, F.: Knowledge neurons in pretrained transformers. arXiv:2104.08696  (2021)

\bibitem{duan2024membership}
Duan, M., Suri, A., Mireshghallah, N., Min, S., Shi, W., Zettlemoyer, L., Tsvetkov, Y., Choi, Y., Evans, D., Hajishirzi, H.: Do membership inference attacks work on large language models? arXiv:2402.07841  (2024)

\bibitem{eldan2023s}
Eldan, R., Russinovich, M.: Who's harry potter? approximate unlearning in llms. arXiv:2310.02238  (2023)

\bibitem{EDPB2024}
{European Data Protection Board}: Opinion 28/2024 on certain data protection aspects related to the processing of personal data in the context of ai models (December 2024), \url{https://www.edpb.europa.eu/system/files/2024-12/edpb_opinion_202428_ai-models_en.pdf}, last accessed on June 2nd, 2025

\bibitem{gao2020making}
Gao, T., Fisch, A., Chen, D.: Making pre-trained language models better few-shot learners. arXiv:2012.15723  (2020)

\bibitem{ginart2019making}
Ginart, A., Guan, M., Valiant, G., Zou, J.Y.: Making ai forget you: Data deletion in machine learning. Advances in neural information processing systems  \textbf{32} (2019)

\bibitem{grattafiori2024llama}
Grattafiori, A., Dubey, A., Jauhri, A., Pandey, A., Kadian, A., Al-Dahle, A., Letman, A., Mathur, A., Schelten, A., Vaughan, A., et~al.: The llama 3 herd of models. arXiv:2407.21783  (2024)

\bibitem{hase2021language}
Hase, P., Diab, M., Celikyilmaz, A., Li, X., Kozareva, Z., Stoyanov, V., Bansal, M., Iyer, S.: Do language models have beliefs? methods for detecting, updating, and visualizing model beliefs. arXiv:2111.13654  (2021)

\bibitem{hayes2025measuring}
Hayes, J., Swanberg, M., Chaudhari, H., Yona, I., Shumailov, I., Nasr, M., Choquette-Choo, C.A., Lee, K., Cooper, A.F.: Measuring memorization in language models via probabilistic extraction. In: Proceedings of the 2025 Conference of the Nations of the Americas Chapter of the Association for Computational Linguistics: Human Language Technologies (Volume 1: Long Papers). pp. 9266--9291 (2025)

\bibitem{huang2022large}
Huang, J., Shao, H., Chang, K.C.C.: Are large pre-trained language models leaking your personal information? arXiv:2205.12628  (2022)

\bibitem{ilharco2022editing}
Ilharco, G., Ribeiro, M.T., Wortsman, M., Gururangan, S., Schmidt, L., Hajishirzi, H., Farhadi, A.: Editing models with task arithmetic. arXiv:2212.04089  (2022)

\bibitem{ippolito2022preventing}
Ippolito, D., Tram{\`e}r, F., Nasr, M., Zhang, C., Jagielski, M., Lee, K., Choquette-Choo, C.A., Carlini, N.: Preventing verbatim memorization in language models gives a false sense of privacy. arXiv:2210.17546  (2022)

\bibitem{jin2024rwku}
Jin, Z., Cao, P., Wang, C., He, Z., Yuan, H., Li, J., Chen, Y., Liu, K., Zhao, J.: Rwku: Benchmarking real-world knowledge unlearning for large language models. arXiv:2406.10890  (2024)

\bibitem{kassem2024alpaca}
Kassem, A.M., Mahmoud, O., Mireshghallah, N., Kim, H., Tsvetkov, Y., Choi, Y., Saad, S., Rana, S.: Alpaca against vicuna: Using llms to uncover memorization of llms. arXiv:2403.04801  (2024)

\bibitem{khattab2023dspy}
Khattab, O., Singhvi, A., Maheshwari, P., Zhang, Z., Santhanam, K., Vardhamanan, S., Haq, S., Sharma, A., Joshi, T.T., Moazam, H., et~al.: Dspy: Compiling declarative language model calls into self-improving pipelines. arXiv:2310.03714  (2023)

\bibitem{kim2023propile}
Kim, S., Yun, S., Lee, H., Gubri, M., Yoon, S., Oh, S.J.: Propile: Probing privacy leakage in large language models. Advances in Neural Information Processing Systems  \textbf{36},  20750--20762 (2023)

\bibitem{lee2022deduplicating}
Lee, K., Ippolito, D., Nystrom, A., Zhang, C., Eck, D., Callison-Burch, C., Carlini, N.: Deduplicating training data makes language models better. In: Proceedings of the 60th Annual Meeting of the Association for Computational Linguistics (Volume 1: Long Papers). pp. 8424--8445 (2022)

\bibitem{li2021large}
Li, X., Tramer, F., Liang, P., Hashimoto, T.: Large language models can be strong differentially private learners. arXiv:2110.05679  (2021)

\bibitem{liu2024large}
Liu, C., Wang, Y., Flanigan, J., Liu, Y.: Large language model unlearning via embedding-corrupted prompts. Advances in Neural Information Processing Systems  \textbf{37},  118198--118266 (2024)

\bibitem{liuRethinkingMachineUnlearning2024a}
Liu, S., Yao, Y., Jia, J., Casper, S., Baracaldo, N., Hase, P., Yao, Y., Liu, C.Y., Xu, X., Li, H., Varshney, K.R., Bansal, M., Koyejo, S., Liu, Y.: Rethinking {{Machine Unlearning}} for {{Large Language Models}} (Dec 2024). \doi{10.48550/arXiv.2402.08787}

\bibitem{liu2025dp}
Liu, Y., Peng, X., Zhang, Y., Ke, X., Deng, S., Cao, J., Ma, C., Fu, M., Zhang, X., Cheng, S., et~al.: Dp-memarc: Differential privacy transfer learning for memory efficient language models. In: Proceedings of the AAAI Conference on Artificial Intelligence. vol.~39, pp. 26317--26325 (2025)

\bibitem{lu2021fantastically}
Lu, Y., Bartolo, M., Moore, A., Riedel, S., Stenetorp, P.: Fantastically ordered prompts and where to find them: Overcoming few-shot prompt order sensitivity. arXiv:2104.08786  (2021)

\bibitem{lucki2409adversarial}
{\L}ucki, J., Wei, B., Huang, Y., Henderson, P., Tramer, F., Rando, J.: An adversarial perspective on machine unlearning for ai safety. URL https://arxiv. org/abs/2409.18025  (2024)

\bibitem{lynch2024eight}
Lynch, A., Guo, P., Ewart, A., Casper, S., Hadfield-Menell, D.: Eight methods to evaluate robust unlearning in llms. arXiv:2402.16835  (2024)

\bibitem{maini2024tofu}
Maini, P., Feng, Z., Schwarzschild, A., Lipton, Z.C., Kolter, J.Z.: Tofu: A task of fictitious unlearning for llms. arXiv:2401.06121  (2024)

\bibitem{meeus2024did}
Meeus, M., Jain, S., Rei, M., de~Montjoye, Y.A.: Did the neurons read your book? document-level membership inference for large language models. In: 33rd USENIX Security Symposium (USENIX Security 24). pp. 2369--2385 (2024)

\bibitem{meng2022locating}
Meng, K., Bau, D., Andonian, A., Belinkov, Y.: Locating and editing factual associations in gpt. Advances in neural information processing systems  \textbf{35},  17359--17372 (2022)

\bibitem{morris2023language}
Morris, J.X., Zhao, W., Chiu, J.T., Shmatikov, V., Rush, A.M.: Language model inversion. arXiv:2311.13647  (2023)

\bibitem{muresanu2024unlearnable}
Muresanu, A., Thudi, A., Zhang, M.R., Papernot, N.: Unlearnable algorithms for in-context learning. arXiv:2402.00751  (2024)

\bibitem{nakka2024pii}
Nakka, K.K., Frikha, A., Mendes, R., Jiang, X., Zhou, X.: Pii-compass: Guiding llm training data extraction prompts towards the target pii via grounding. arXiv:2407.02943  (2024)

\bibitem{nasr2023scalable}
Nasr, M., Carlini, N., Hayase, J., Jagielski, M., Cooper, A.F., Ippolito, D., Choquette-Choo, C.A., Wallace, E., Tram{\`e}r, F., Lee, K.: Scalable extraction of training data from (production) language models. arXiv:2311.17035  (2023)

\bibitem{nguyen2022survey}
Nguyen, T.T., Huynh, T.T., Ren, Z., Nguyen, P.L., Liew, A.W.C., Yin, H., Nguyen, Q.V.H.: A survey of machine unlearning. arXiv:2209.02299  (2022)

\bibitem{noyb2024chatgpt}
{NOYB}: Chatgpt provides false information about people, and openai can’t correct it. \url{https://noyb.eu/en/chatgpt-provides-false-information-about-people-and-openai-cant-correct-it} (2024), accessed: 2025-06-02

\bibitem{pandaPrivacyAuditingLarge2025}
Panda, A., Tang, X., Nasr, M., {Choquette-Choo}, C.A., Mittal, P.: Privacy {{Auditing}} of {{Large Language Models}} (Mar 2025). \doi{10.48550/arXiv.2503.06808}

\bibitem{patil2023can}
Patil, V., Hase, P., Bansal, M.: Can sensitive information be deleted from llms? objectives for defending against extraction attacks. arXiv:2309.17410  (2023)

\bibitem{pawelczyk2023context}
Pawelczyk, M., Neel, S., Lakkaraju, H.: In-context unlearning: Language models as few shot unlearners. arXiv:2310.07579  (2023)

\bibitem{puerto2024scaling}
Puerto, H., Gubri, M., Yun, S., Oh, S.J.: Scaling up membership inference: When and how attacks succeed on large language models. arXiv:2411.00154  (2024)

\bibitem{schlangen-2021-targeting}
Schlangen, D.: Targeting the benchmark: On methodology in current natural language processing research. arXiv:2007.04792  (2020)

\bibitem{NEURIPS2021_9627c45d}
Sekhari, A., Acharya, J., Kamath, G., Suresh, A.T.: Remember what you want to forget: Algorithms for machine unlearning. In: Ranzato, M., Beygelzimer, A., Dauphin, Y., Liang, P., Vaughan, J.W. (eds.) Advances in Neural Information Processing Systems. vol.~34, pp. 18075--18086. Curran Associates, Inc. (2021)

\bibitem{shani2025tokens}
Shani, C., Jurafsky, D., LeCun, Y., Shwartz-Ziv, R.: From tokens to thoughts: How llms and humans trade compression for meaning. arXiv:2505.17117  (2025)

\bibitem{shi2023detecting}
Shi, W., Ajith, A., Xia, M., Huang, Y., Liu, D., Blevins, T., Chen, D., Zettlemoyer, L.: Detecting pretraining data from large language models. arXiv:2310.16789  (2023)

\bibitem{shin2020autoprompt}
Shin, T., Razeghi, Y., Logan~IV, R.L., Wallace, E., Singh, S.: Autoprompt: Eliciting knowledge from language models with automatically generated prompts. arXiv:2010.15980  (2020)

\bibitem{staab2023beyond}
Staab, R., Vero, M., Balunovi{\'c}, M., Vechev, M.: Beyond memorization: Violating privacy via inference with large language models. arXiv:2310.07298  (2023)

\bibitem{team2024gemini}
Team, G., Georgiev, P., Lei, V.I., Burnell, R., Bai, L., Gulati, A., Tanzer, G., Vincent, D., Pan, Z., Wang, S., et~al.: Gemini 1.5: Unlocking multimodal understanding across millions of tokens of context. arXiv:2403.05530  (2024)

\bibitem{thaker2024guardrail}
Thaker, P., Maurya, Y., Hu, S., Wu, Z.S., Smith, V.: Guardrail baselines for unlearning in llms. arXiv:2403.03329  (2024)

\bibitem{tiwari2024sequence}
Tiwari, T., Suh, G.E.: Sequence-level analysis of leakage risk of training data in large language models. arXiv:2412.11302  (2024)

\bibitem{pmlr-v134-ullah21a}
Ullah, E., Mai, T., Rao, A., Rossi, R.A., Arora, R.: Machine unlearning via algorithmic stability. In: Belkin, M., Kpotufe, S. (eds.) Proceedings of Thirty Fourth Conference on Learning Theory. Proceedings of Machine Learning Research, vol.~134, pp. 4126--4142. PMLR (15--19 Aug 2021)

\bibitem{ulmer2024uncertainty}
Ulmer, D.: On uncertainty in natural language processing. arXiv:2410.03446  (2024)

\bibitem{vilellaDeIndexingRightBe2025}
Vilella, S., Ruffo, G.: ({{De}})-{{Indexing}} and the {{Right}} to be {{Forgotten}} (Jan 2025). \doi{10.48550/arXiv.2501.03989}

\bibitem{wang2025selective}
Wang, L., Zeng, X., Guo, J., Wong, K.F., Gottlob, G.: Selective forgetting: Advancing machine unlearning techniques and evaluation in language models. In: Proceedings of the AAAI Conference on Artificial Intelligence. vol.~39, pp. 843--851 (2025)

\bibitem{yao2024machine}
Yao, J., Chien, E., Du, M., Niu, X., Wang, T., Cheng, Z., Yue, X.: Machine unlearning of pre-trained large language models. In: Proceedings of the 62nd Annual Meeting of the Association for Computational Linguistics (Volume 1: Long Papers). pp. 8403--8419 (2024)

\bibitem{yao2024large}
Yao, Y., Xu, X., Liu, Y.: Large language model unlearning. Advances in Neural Information Processing Systems  \textbf{37},  105425--105475 (2024)

\bibitem{zhang2023counterfactual}
Zhang, C., Ippolito, D., Lee, K., Jagielski, M., Tram{\`e}r, F., Carlini, N.: Counterfactual memorization in neural language models. Advances in Neural Information Processing Systems  \textbf{36},  39321--39362 (2023)

\bibitem{zhangRightBeForgotten2024}
Zhang, D., {Finckenberg-Broman}, P., Hoang, T., Pan, S., Xing, Z., Staples, M., Xu, X.: Right to be forgotten in the {{Era}} of large language models: Implications, challenges, and solutions. AI and Ethics  (Sep 2024)

\bibitem{zhouQuantifyingAnalyzingEntityLevel2024}
Zhou, Z., Xiang, J., Chen, C., Su, S.: Quantifying and {{Analyzing Entity-Level Memorization}} in {{Large Language Models}}. Proceedings of the AAAI Conference on AI  \textbf{38}(17),  19741--19749 (Mar 2024). \doi{10.1609/aaai.v38i17.29948}

\end{thebibliography}

\end{document}